# Open and Sustainable AI: challenges, opportunities and the road ahead in the life sciences

Gavin Farrell[1], Eleni Adamidi[2], Rafael Andrade Buono[3], Mihail Anton[4], Omar Abdelghani Attafi[1], Salvador Capella Gutierrez[5], Emidio Capriotti[6,7], Leyla Jael Castro[8], Davide Cirillo[5], Lisa Crossman[9,10], Christophe Dessimoz[11,12], Alexandros Dimopoulos[13,14], Raúl Fernández-Díaz[15,16,17], Styliani-Christina Fragkouli[18,19], Carole Goble[20], Wei Gu[21], John M. Hancock[22], Alireza Khanteymoori[23], Tom Lenaerts[24,25,26,27,28], Fabio G. Liberante[4], Peter Maccallum[4], Alexander Miguel Monzon[1], Magnus Palmblad[29], Lucy Poveda[12], Ovidiu Radulescu[30], Denis C. Shields[15,16], Shoaib Sufi[20], Thanasis Vergoulis[2], Fotis Psomopoulos[18,*], Silvio C.E. Tosatto[1,31,*]

***Co-corresponding authors**: [18]Institute of Applied Biosciences, Centre for Research and Technology Hellas, Thessaloniki, Greece and [1]Department of Biomedical Sciences, University of Padova, Padova, 35121, Italy. **E-mails**: fpsom@certh.gr, silvio.tosatto@unipd.it

[1] Department of Biomedical Sciences, University of Padova, Via Ugo Bassi 58/B, 35131 Padova, Italy
[2] Athena Research & Innovation Center, Aigialias & Chalepa, 15125, Marousi, Greece
[3] VIB Data Core, VIB.AI Center for AI & Computational Biology, Technologiepark-Zwijnaarde 75, 9052 Ghent, Belgium
[4] ELIXIR Europe Hub, EMBL-EBI South Building, Wellcome Genome Campus, Hinxton CB10 1SD, United Kingdom
[5] Barcelona Supercomputing Center (BSC), Plaça Eusebi Güell, 1-3, 08034, Barcelona, Spain
[6] Department of Pharmacy and Biotechnology, University of Bologna, 40126 Bologna, Italy
[7] Computational Genomics Platform, IRCCS University Hospital of Bologna, 40138 Bologna, Italy
[8] ZB MED Information Centre for Life Sciences, Gleueler Straße 60, 50931 Cologne, Germany
[9] SequenceAnalysis.co.uk, United Kingdom
[10] University of East Anglia, Research Park, Norwich NR4 7TJ, United Kingdom
[11] Department of Computational Biology, University of Lausanne, Genopode 2024.3, 1015 Lausanne, Switzerland
[12] Swiss Institute of Bioinformatics, Amphipôle, Quartier UNIL-Sorge, 1015 Lausanne, Switzerland
[13] Institute for Fundamental Biomedical Science, Biomedical Sciences Research Center 'Alexander Fleming', Vari, Greece
[14] Department of Informatics & Telematics, School of Digital Technology, Harokopio University, Athens, Greece
[15] School of Medicine, University College Dublin, Belfield, Dublin 4, Ireland
[16] Conway Institute of Biomolecular and Biomedical Research, University College Dublin, Belfield, Dublin 4, Ireland

[17] IBM Research Dublin, Dublin D15 HN66, Ireland
[18] Institute of Applied Biosciences, Centre for Research and Technology Hellas, Thessaloniki, 57001, Greece
[19] Department of Biology, National & Kapodistrian University of Athens, Athens, 15772, Greece
[20] Department of Computer Science, University of Manchester, Oxford Road, Manchester, M19 3PL, United Kingdom
[21] Luxembourg National Data Service, 6, avenue des Hauts-Fourneaux, L-4362 Esch-sur-Alzette, Luxembourg
[22] Institute of Biochemistry and Molecular Genetics, Faculty of Medicine, University of Ljubljana, 1000 Ljubljana, Slovenia
[23] Department of Psychology, University of Freiburg, Freiburg, Germany
[24] Machine Learning Group, Université Libre de Bruxelles, Brussels, Belgium
[25] Artificial Intelligence Lab, Vrije Universiteit Brussel, Brussels, Belgium
[26] Interuniversity Institute of Bioinformatics in Brussels, ULB-VUB, Brussels, Belgium
[27] FARI, AI for the common good institute, ULB-VUB, Brussels, Belgium
[28] Center for Human-Compatible AI, UC Berkeley, Berkeley, CA, USA
[29] Leiden University Medical Center, Leiden 2333 ZG, Netherlands
[30] LPHI, University of Montpellier, CNRS, INSERM, Montpellier, France
[31] Institute of Biomembranes, Bioenergetics and Molecular Biotechnologies, National Research Council (CNR-IBIOM), Bari 70126, Italy

# Abstract

Artificial intelligence (AI) has recently seen transformative breakthroughs in the life sciences, expanding possibilities for researchers to interpret biological information at an unprecedented capacity, with novel applications and advances being made almost daily. In order to maximise return on the growing investments in AI-based life science research and accelerate this progress, it has become urgent to address the exacerbation of long-standing research challenges arising from the rapid adoption of AI methods. We review the increased erosion of trust in AI research outputs, driven by the issues of poor reusability and reproducibility, and highlight their consequent impact on environmental sustainability. Furthermore, we discuss the fragmented components of the AI ecosystem and lack of guiding pathways to best support Open and Sustainable AI (OSAI) model development. In response, this perspective introduces a practical set of OSAI recommendations directly mapped to over 300 components of the AI ecosystem and provides guiding implementation pathways. Our work connects researchers with relevant AI resources, facilitating the implementation of sustainable, reusable and reproducible AI. Built upon life science community consensus and aligned to existing efforts, the outputs of this perspective are designed to aid the future development of policy and additional structured pathways for guiding AI implementation.

# Introduction

The field of artificial intelligence (AI) has seen a surge in breakthrough advances across scientific domains where researchers are developing AI models at an unprecedented rate, a trend directly reflected in the proliferation of AI-focused publications[1]. In particular, the life sciences domain has seen a rapid adoption and integration of AI models, spurred on by an increasing volume of expertly curated biological data through the FAIR movement[2] and democratised access to powerful hardware such as Graphical Processing Units (GPU). Additionally, landmark successes like AlphaFold 2[3], which addressed the decades-old challenge of protein structure prediction, are leading to a snowballing effect in AI adoption[4]. This is evidenced by AI model applications upending traditional biological computational approaches and being integrated into life science databases and tooling[5,6].

In spite of this seemingly positive openness to adopt AI in the life sciences, there is a darker, publication-rooted AI methodology crisis[1,7], where many recent AI publications are of limited reuse value due to their opaque model descriptions[8]. These failings frequently stem from poor disclosure and provenance of the data, code, and model details, leading to challenges in reusability and reproducibility, such as inaccessible data[9], incomplete code repositories[10], or missing model information. Researchers outputting these inadequately described publications may be unaware of the importance of these elements, an issue compounded by a lack of dedicated guidance to help researchers action AI model sharing best practices. Furthermore, insufficient recognition for extra efforts to comprehensively describe models, and a low requirement by journals for detailed AI method disclosures act as additional catalysts to the proliferation of poorly described AI publications.

To address these systemic issues, lessons can be drawn from successful research practice reforms. The FAIR data movement[2] is a key example of reform in the life sciences, and was in part supported in its wide application through community-developed toolkits (e.g., Research Data Management Toolkit[11], Pistoia Alliance Toolkit[12]). These toolkit resources act as central hubs of community maintained knowledge and can host guided implementation pathways for research best practices. In contrast, AI researchers in the life sciences currently lack such dedicated toolkits and face challenges in navigating the fragmented ecosystem of AI components like model registries[13,14] and metadata standards[15,16]. Without guiding AI best practice implementation pathways, progress towards widely available reusable and reproducible models will be impeded.

A less visible issue magnified by the low reusability and reproducibility of AI models is environmental sustainability[17]. Where a model cannot be reused or extensive experimentation is needed to reproduce it, wasteful resource usage occurs. This, paired with the growth in life scientists adopting more resource-intensive deep learning approaches like biological large language models (LLMs)[18] and agentic AI[19], will exacerbate environmental impacts if not addressed. These issues, compounded by a lack of Green AI[20] development techniques and rare reporting of environmental impacts, demand attention to ensure accountable and sustainable AI in the life sciences.

It has therefore become urgent to anticipate and mitigate the growing risks associated with insufficient descriptions of AI methodologies, which spur poor reuse, reproducibility and impact environmental sustainability. This perspective highlights these risks and challenges from the life sciences point of view, taking into consideration existing global efforts (e.g., Bridge2AI[8], Model Openness Framework[21], FUTURE-AI[22], REFORMS[23], MLCommons[24], FARR-RCN[25]). The scope of our work addresses key challenges to achieving Open and Sustainable AI (OSAI) model development in the life sciences through improving reuse, reproducibility, and environmental sustainability. We propose an accessible set of guiding OSAI recommendations, with novelty in their mapping to a collection of over 300 AI components that can support their implementation alongside three reference implementation pathways. While acknowledging the wider scope of AI research challenges, this perspective focuses on areas where there are available solutions to support a collective and international implementation. Therefore, we do not directly address the areas of Explainable AI[26], Trustworthy AI[27], Ethical AI[28] or the wider set of topics within Responsible AI[29], which raise complex socio-ethical considerations and often require more nuanced approaches. Our goal is to guide and connect researchers with solutions to enhance sustainable and collaborative AI-based research, accelerating life science discovery.

# Current Challenges

## Reusability of AI models

Despite the rapid growth in publications of new life science AI methods[1], they face a major lack of reuse such as through adaptation and redeployment. Even within the subset of life science models that do see reuse, there are obstacles to their full exploitation. A notable example is AlphaFold 2[3], which was missing crucial model information. This impeded its full reuse potential, which necessitated a more open community reimplementation, OpenFold[30], that shared all model relevant code and data. The barriers to reuse of AI models in the life sciences stem from a diverse number of reasons such as: narrow task applicability; the academic requirement for novelty; inequitable resource access; and the lack of explainability and trustworthiness of deep learning based models. While not all these aspects impacting reuse can be addressed, there are a number of more practical issues that can be addressed through researcher best practice reform (Fig. 1). These include absent or restrictive licensing, insufficient metadata, and low rates of dedicated AI registry usage.

**Figure 1: Barriers to Open and Sustainable AI in the life sciences**

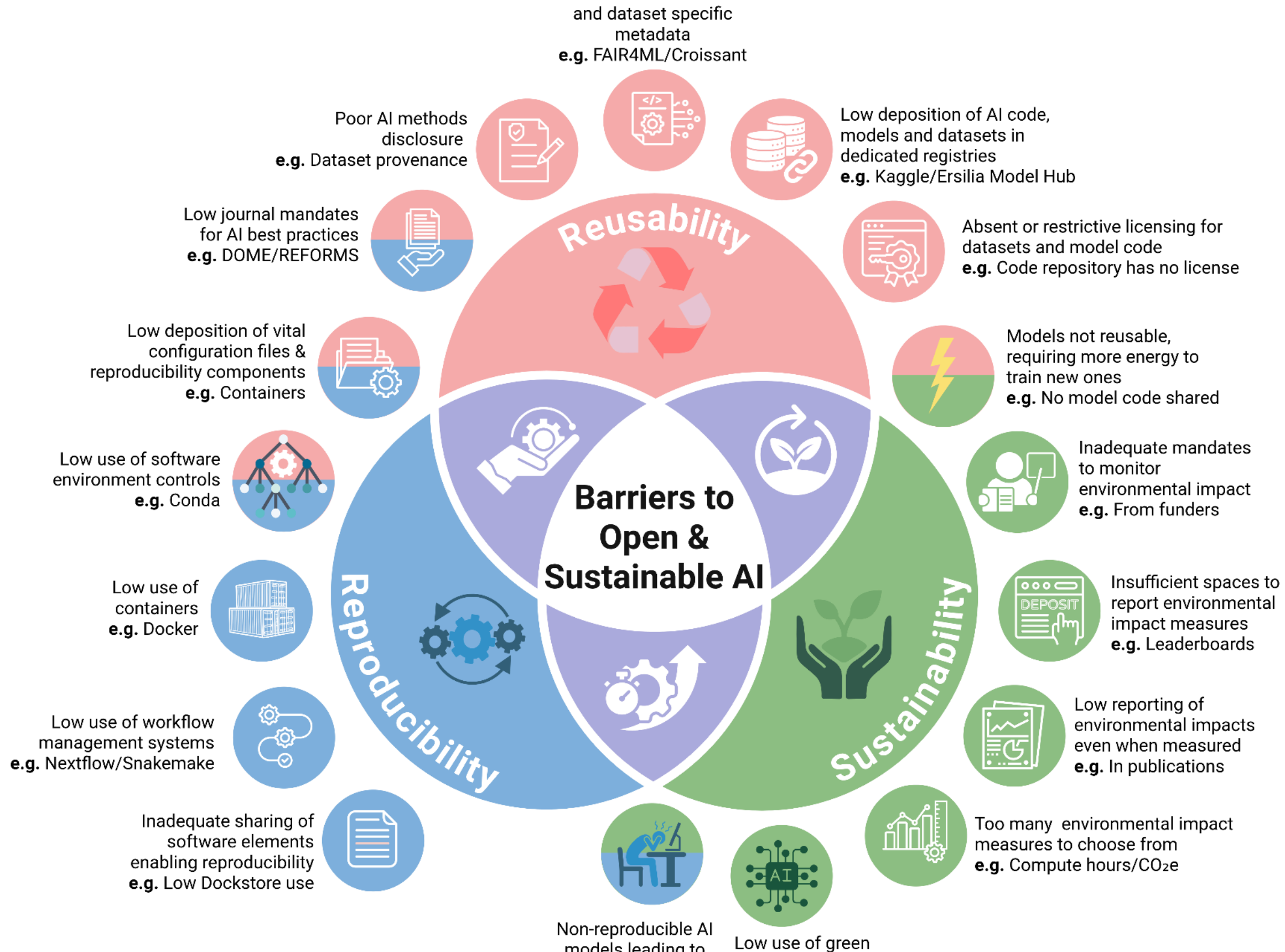

Three interlinked challenge areas of reusability, reproducibility and sustainability that must be overcome by researchers to accelerate Open and environmentally sustainable AI model development in the life sciences. Challenge areas have their key issues highlighted with examples, as well as their intersections.

There are many existing community developed recommendations and best practices that can be leveraged to improve AI model publishing such as the DOME Recommendations[1], TRIPOD[31], REFORMS[23], and Bridge2AI[8]. Additionally, there are several life science sub-domain checklists such as Schmied et al.'s (2023) checklist for AI in bioimaging[32] and CONSORT-AI[33]/SPIRIT-AI[34] designed for AI use in clinical trials. These resources can aid a researcher to correctly share all relevant model and dataset information. Unfortunately, many of these face low adoption amongst researchers and this issue is compounded by the loose or lack of journal mandates for AI specific methods disclosures in publications. For publishers, their inclusions for AI model specific software and data are generally loose beyond standard life science database and public software repository links such asGitHub or GitLab. During peer review, dataset and code repository links a researcher provides can often go ignored, with a reviewer instead solely focusing on the publication text. The current emphasis in research publishing involves text-based descriptions taking priority over a comprehensive sharing of the methods for real-world reuse. This leaves the practical considerations for supporting AI model reuse as an afterthought, if they are addressed at all. There are also few motivating incentives to encourage a researcher to take these extra steps despite their value to scientific advancements.

Researchers are also currently faced with too many options and a functionally diverse AI ecosystem of repositories to document and store their AI datasets, code and models. These include AI domain-agnostic solutions (e.g., Kaggle[35], Hugging Face[36]), life science specific resources (e.g., Kipoi[14], BioImage Model Zoo[13], Ersilia Model Hub[37]), or even trusted non-AI platforms (e.g., Zenodo[38], in structured RO-Crates[39]). This choice forces consideration of the diverse feature sets, user bases, long-term data retention policies, and trustworthiness of different platforms. As a result, this has often led to inconsistent sharing, poor findability, and models not being deposited in a reusable manner beyond basic raw file availability. Furthermore, a lack of trust in a potentially suitable method, due to insufficient information to fully assess it, often hinders reuse and encourages the development of new models rather than reimplementing potentially suitable ones. This can be overcome by addressing the failings of proper technical sharing of AI data, code, models, and their documentation covering important aspects such as their dataset availability and provenance[39].

FAIR for AI initiatives[40] are looking to support reusability globally across both academia and industry, with organisations like the Research Data Alliance (RDA) FAIR4ML Interest Group[16] developing model metadata (FAIR4ML schema[41], Machine Learning Commons (MLCommons)[24] focusing on metadata for model datasets (Croissant[15]), and the Pistoia Alliance[42] addressing AI-data readiness[43] (DataFAIRy project[44]). However, in spite of this initial work, widely available and truly reusable AI models are still not yet prevalent. This is due to a multitude of reasons including the rapid increase in AI adoption outpacing best practice development, and critically, a lack of targeted top-down best practice mandates from funders, policy makers, and journals.

## Reproducibility of AI models

Reproducibility is a multifaceted concept and in the context of AI research has various interpretations[45]. We describe the challenge of AI model reproducibility in relation to the reuse of a method's available data and code to: facilitate a reproduction of an AI model to enhance understanding; and reproduce consistent model benchmarking results to build trust in reported performances. These are vital to extract learnings and validate previous findings from promising research to build upon with future work[46]. Highlighting the widespread difficulty of AI model reproducibility in the life sciences, BioModelsML attempted to curate a collection of reproducible models from public sources but faced many time-consuming barriers during their efforts[47]. Even in cases where all relevant information towards generating an AI model is openly available and clearly disclosed, software environment requirements can often be prohibitively time-consuming and impractical to replicate. Therefore, an AI model can be transparently described, but practically not reproducible due to poor portability. As such, the adoption of solutions that can support standardised reproducibility of AI methods across systems requires a broader uptake to address the reproducibility crisis facing researchers.

On the software side, several well-established resources exist to facilitate AI model portability and computational reproducibility. Many of these solutions span the areas of software containerisation (e.g., Docker[48]), environment control (e.g., Conda[49]), and workflow management (e.g., Nextflow[50], Snakemake[51], Galaxy[52]). These solutions supporting portable deployment across systems, provide simpler reproductions of complex computational runs with far less overheads[53]. While these solutions help reduce the barriers to life science computing reproducibility, their prevalence and use for AI models is still variable amongst publications[47]. Even though many of these reproducibility-enabling tools cover AI needs, many AI methods only share a disjoint repository of scripts with limited instructions to easily reproduce their entire runs. Therefore, AI model reproducibility is time-consuming due to the infrequent and inconsistent application of tools capable of orchestrating complex model generation runs. Reasons for their underutilisation may include the time and training needed to leverage them, as well as a misunderstanding by researchers of the overheads involved to reproduce an AI model across diverse computational environments.

Beyond software approaches for ensuring reproducibility of AI methods, the aspect of diverse physical hardware must also be considered. Computational infrastructure requirements can range drastically in developing and deploying AI models, from use of a simple laptop CPU or Google Colab[54] instance to requiring a cluster using hundreds of GPUs. Consequently, correct documentation of compute infrastructure used and configuration file sharing is important for reproducibility[45] but often lacking in life science AI model sharing. This oversight means that attempting to run and reproduce an AI method may be impossible, even when access to the required compute infrastructure is not a limitation. Moreover, researchers can face prohibitive costs when using commercial cloud solutions in attempts to reproduce an AI model where the exact hardware needs of the original model were not openly disclosed. This creates unforeseen access challenges for researchers wishing to reproduce and build upon existing AI methods. Therefore, it is crucial for researchers to closely consider transparently documenting all infrastructure and related configuration requirements in addition to taking software measures to provide portable reproducibility.

## Sustainability of AI models

The challenges of poor AI method reusability and reproducibility directly impact environmental sustainability, due to the need to repeat heavy energy expenditures such as retraining models, which increases carbon emissions[17]. This "Red AI"[20] issue is compounded by the escalating resource needs of modern AI systems, particularly deep learning models. Their adoption by a wider number of life scientists intensifies the growing energy demands[55] and raises Responsible AI concerns about the global north's disproportionate environmental impact[56]. Furthermore, where models are not made available and preserved, this adds to irresponsible AI energy waste. Therefore, environmental sustainability of AI systems requires accountable model development in ways that minimise negative environmental impacts while maximising societal outcomes, a principle reflected in initiatives like the European Green Deal[57], which closely considers AI impacts.

Despite these Red AI concerns, there is not yet a prominent focus in life science research on mitigating it through the adoption of Green AI[58] approaches. This is partly because applying AI to biological data is already an inherently challenging and interdisciplinary task, often leaving less pressing Green AI considerations like measuring and reporting environmental sustainability impacts a lower priority. The implementation and routine reporting of energy usage metrics, especially outside large-scale HPC sites, is not standard practice. This lack of monitoring and reporting often stems from insufficient requirements by organisations or funders, as AI sustainability is frequently a less visible concern. While individual projects may seem to have a small footprint, the collective global growth in the adoption of energy-intensive deep learning represents a largely unmonitored aggregate impact, necessitating better tracking. Even where large HPC centres track and report their overall consumption (e.g., EMBL-EBI energy consumption report[59]), environmental impact accounting attributable to specific AI model development often remains unaddressed at the individual lab and researcher level.

Compounding these issues, even when individual researchers are conscious of AI sustainability impacts, they may find it difficult to know what to measure, in what format, and where to report it. Consequently, they may face challenges in identifying and understanding the limitations of common sustainability measures like $CO_2$ equivalent or energy consumption, each varying in accuracy and applicability (Table 1). This situation is exacerbated by a lack of accessible tooling designed for life science researchers to correctly plan, track, and measure the sustainability impact of their AI methods. Furthermore, for reporting measures there is an absence of journal requirements and reporting templates.

**Table 1: Metrics a researcher could use to measure AI model computation and energy usage to infer environmental impact.**

| Measure | Description | Measurement limitation | Unit |
| --- | --- | --- | --- |
| $CO_2e$ ($CO_2$ equivalent) | Estimated greenhouse gas emissions associated with the energy consumed, based on grid carbon intensity. | An estimate, highly dependent on location, time, and accuracy of the energy measurement and carbon intensity factor used. Doesn't include hardware's embodied carbon. | Kilograms $CO_2e$ (kg $CO_2e$)<br><br>Tonnes $CO_2e$ (t $CO_2e$) |

| Energy consumption | Total electrical energy consumed by the hardware during the computation. Direct measure of resource use. | Can be difficult to measure accurately without dedicated power meters. Software estimates may be inaccurate or incomplete (do not account for RAM, cooling, etc.). | Watt-hours (Wh)<br><br>Joules (J) |
|---|---|---|---|
| Floating-point operations | Count of arithmetic operations on floating-point numbers; proxy for algorithmic complexity. | Hardware-independent, doesn't directly measure time or energy. Ignores data movement costs (memory input output). Efficiency (FLOPs/Watt or FLOPs/second) varies vastly across hardware. | FLOPs (e.g., Giga-FLOP) |
| GPU/CPU hours | Total time specific compute units (GPU cores, CPU cores) were allocated or actively used for the task. | Doesn't account for varying utilisation levels (idle vs. 100%). Different hardware has vastly different power consumption per hour. Poor indicator of actual energy used. | Hours (h)<br><br>core-hours (e.g., GPU-hours) |
| Power draw | Instantaneous rate at which electrical energy is consumed by the hardware components. | Fluctuates significantly during computation. Average power needed for energy calculation. Thermal design power is often a limit, not actual usage. Software estimates vary in accuracy. | Watts (W)<br><br>kilowatts (kW) |
| Total runtime | Wall-clock time elapsed from the start to the end of the computation process. | Highly dependent on specific hardware used. Doesn't distinguish computation vs. idle/wait time. Doesn't directly measure energy or computational work performed. | Seconds (s)<br><br>Minutes (min)<br><br>Hours (h) |

Beyond measuring and reporting, a critical implementation gap exists in the life sciences concerning the adoption of methods to reduce AI's environmental impact, such as through Green AI model development. Many approaches exist such as Frugal AI[60], which considers energy efficient model complexity reduction, and Green AutoML techniques[61] for optimising architecture while minimising computational cost. However, such techniques remain underutilised, leading to inefficient resource use.

Overall, while wider Responsible AI governance is visible in the life sciences, often with a strong Ethical, Legal and Societal Implications (ELSI)[62] emphasis on human data privacy (e.g., GRAIMatter highlighting AI privacy leakage risks[63]), specific mandates for AI environmental sustainability are frequently overlooked when not driven by institutional and governing bodies. A possible factor in this context may include the ethical dilemma of attempting to balance conscious energy use against vital research often justified for societal good like cancer studies or drug discovery. This highlights the need to carefully manage the relation between important research freedom alongside mitigating environmental sustainability concerns. Encouragingly, initiatives like the FUTURE-AI consortium[22] and the AHEAD project[64] are beginning to develop broader assessment frameworks applicable to AI in the life sciences that include technical, clinical, socio-ethical, and legal dimensions, which may pave the way for more holistic environmental sustainability evaluations. Furthermore, emerging project initiatives considering Green AI (e.g., in the ELIXIR STEERS project[65], SustAInML EU project[66]) or community driven sustainable software accreditations (e.g., Green DiSC Certification[67]) signal a positive direction for integrating environmentally conscious solutions to meet urgently needed Green AI needs in the life sciences.

# Open and Sustainable AI: Recommendations

In direct response to the three pressing and interlinked challenges of reusability, reproducibility and environmental sustainability, we propose nine practical and achievable OSAI recommendations that life science researchers should implement. These recommendations are structured in three tiers of implementation priority and designed to address the most overlooked but impactful considerations to creating OSAI compliant research outputs. We recognise the pre-existing body of work to guide researchers leveraging AI such as Bridge2AI[8] for biomedical research AI-data readiness and the domain-agnostic REFORMS[23] AI disclosure checklist. However, our recommendations offer a targeted and implementation oriented approach for the broader life sciences, specifically addressing the urgent challenges concerning both AI models and their datasets that are prevalent across the entire domain. We provide novelty with our recommendations through their support for application with a mapping back to a non-prescriptive collection of over 300 AI ecosystem components, such as AI specific metadata, registries and tools (https://osai.dome-ml.org/ai-ecosystem). This ensures researchers are both informed of the best practice recommendations and linked with the existing solutions to accomplish OSAI.

## Reusability

**R1: Generate and share standardised AI metadata for models and datasets**

Standardised metadata is essential to aid both human and machine understanding of AI models and their datasets for reuse. AI metadata adoption enables programmatic findability, supports interoperability, and ensures correct disclosure of key model and dataset attributes. We recommend generating and exposing metadata for both AI models and their respective datasets. For model annotation, a possible standard includes the recently launched FAIR4ML schema[41] from the RDA FAIR4ML Interest Group[16]. For AI dataset annotation, the Croissant[15] standard from MLCommons[24] is a suitable example due to its layered approach and domain extensibility (e.g., GeoCroissant for geospatial AI datasets[68]). Both are open, community-driven standards and cover diverse metadata fields such as key information like licensing and provenance. Other AI metadata standards also exist such as Model Cards[69] and Dataset Cards[70]. Additional considerations include leveraging dedicated AI ontologies like Machine Learning Sailor Ontology (MLSO)[71] to help standardise metadata vocabulary.

**R2: Leverage AI registries as central hubs for sharing and discovering reusable AI models and datasets**

There is a growing number of AI model and dataset registries, with over 30 identified in our AI ecosystem landscaping. They support depositing, registering and sharing these valuable research objects for community reuse. While choosing the most suitable one can be difficult, a wider adoption by researchers will enhance visibility, aid co-development and functionality convergence to serve life science needs. The primary choice consideration relates to choosing either a suitable life science specific registries (e.g., BioImage Model Zoo[13] for bioimaging, Kipoi[14] for genomics) or a domain-agnostic registries (e.g., Hugging Face[36], Kaggle[35]). Additional considerations supporting this choice include the AI registry provider's commitment to Open Science, long-term data preservation, and support for exposing standardised model and dataset metadata.

**R3: Host, promote and share training and guidance on correct deposition of AI research objects for reuse**

There are many evolving AI model deposition considerations such as metadata, hosting registries and related checklists, making it difficult for researchers to keep up with best practices to support AI model and dataset sharing. It is therefore crucial that research groups and institutions begin to develop, host and promote training and guidance on correct AI model deposition. This will aid researchers to comprehensively share models and maximise reuse value. While AI model deposition training material examples are limited, there are institutional model deposition guidance examples such as the Swedish SciLifeLab lightweight and nationally focused guidance for model sharing[72], which reuses existing frameworks such as the Model Openness Framework[21]. Furthermore, to democratise the availability of these materials, we recommend hosting them in existing open infrastructure like training material hosting resources (e.g., Glittr[73], Data Carpentry[74]) or Turing Way[75] modelled toolkits (e.g., RDMkit[11]). This will be essential to democratise model sharing knowledge and maximise reusability.

**Box 1. Recommendations to aid AI model reuse**

| **Reusability** | | | |
|---|---|---|---|
| **No.** | **Recommendation** | **Explanation** | **OSAI Priority** |
| R1 | Generate and share standardised AI metadata for models and datasets | Adopt and share standardised metadata using community-driven schemas for both datasets and models to enhance machine readability, findability, and interoperability for reuse. | Essential |
| R2 | Leverage AI registries as central hubs for sharing and discovering reusable AI models and datasets | Deposit and register models and datasets in established AI registries, considering choice between domain-specific or general-purpose options to maximise visibility and accessibility for the research community. | Required |
| R3 | Host, promote and share training and guidance on correct deposition of AI research objects for reuse | Develop and disseminate training materials on best practices for AI model and dataset deposition, utilising dedicated training resource platforms to build community capacity and ensure proper sharing for reuse. | Recommended |

## Reproducibility

**R4: Ensure transparent disclosure, clear documentation and sharing of all model relevant information**

Transparent reporting of all model-relevant information is crucial to facilitate understanding and reproducibility, even if complete computational reproducibility is challenging. This includes sharing comprehensive details on model architecture and weights, preprocessing steps, model architecture, training procedures, datasets, evaluation metrics, and ideally complemented by access to relevant reproducibility files (e.g., ONNX[76]). Such information should be readily available across a publication's methods sections, supplementary materials, or alongside code in repositories. These disclosures should be guided by the use of comprehensive checklists (e.g., Model Openness Framework[21], REFORMS[23]) and where possible supported by their hosting in reporting platforms like the DOME Registry[77], which provides structured information access and persistent identifiers for this information. Ultimately transparent disclosure is crucial for ensuring clear access to information to aid reproducibility.

**R5: Provide portable code and reproducible environments to facilitate smooth reuse**

To enable practical reproduction of models from training to inference across diverse computational systems, researchers should leverage readily available components. These include using software containers (e.g., Docker[48], Apptainer[78]), environment control (e.g., Conda[49]), and workflow management systems (e.g., Nextflow[50], Snakemake[51]). These should then be made available for use through open platforms (e.g., DockerHub[79], Dockstore[80]). While simpler methods like online and interactively hosted scripts (e.g., Google Colab[54], Code Ocean Capsules[81]) can be useful, sharing dedicated reproducibility components within well-annotated code repositories alongside all necessary configuration files represents best practice. Choice of tooling should consider longevity and support for components used to ensure long-term reproducibility.

**R6: Use standardised AI-ready datasets and benchmarking evaluation protocols to facilitate reproducible model comparisons**

To ensure the validity of model performance claims and enable reproducible comparisons, the use of standardised evaluation benchmark datasets and protocols across models is required. AI-ready datasets as defined by the Open Data Institute's (ODI) 'Framework for AI-ready data'[43], are those which are structured and prepared for direct use by AI models across areas such as technical optimisation and adherence to standards. They are key to facilitating reproducible and comparable evaluations. Additionally, rigour in standardising model evaluation protocols is vital for transparently reporting model performances. Community-driven efforts, such as challenges like CASP[82] for protein structure prediction or wider collection of DREAM challenges[83] (e.g., drug sensitivity prediction) are core to agreeing and applying standardised evaluation protocols. These efforts in turn drive scientific discovery and the development of more trustworthy and reliable model assessments. Platforms like OpenEBench[84] (life science specific) and Kaggle competitions[35] (domain-agnostic) can aid in discovering and hosting benchmarking challenge results and information on the standardised datasets, and evaluation protocols enabling reliable comparisons.

**Box 2. Recommendations to aid AI model reproducibility**

| **Reproducibility** | | | |
|---|---|---|---|
| **No.** | **Recommendation** | **Explanation** | **OSAI Priority** |
| R4 | Ensure transparent disclosure, clear documentation and sharing of all model relevant information | Ensure transparent reporting by comprehensively documenting all model-relevant information, including data, methods, and configurations, guided by community disclosure checklists and reporting platforms. | Essential |
| R5 | Provide portable code and reproducible environments to facilitate smooth reuse | Facilitate practical model execution by providing portable code within reproducible environments using software containerisation, environment management tools, and workflow systems. | Required |

| R6 | Use standardised AI-ready datasets and benchmarking evaluation protocols to facilitate reproducible model comparisons | Promote model comparability and validate performance claims by using standardised benchmark datasets and community-agreed evaluation protocols, often accessible via public benchmarking platforms | Recommended |
|---|---|---|---|

## Sustainability

### R7: Implement Green AI model development techniques

The diverse set of Green AI model development techniques can greatly reduce energy consumption during model training and boost downstream inference efficiency. Methods such as warmstarting[85], pruning[86], quantisation[87], green auto-ML[88], knowledge distillation[88], and transfer learning[89] can significantly cut compute needs. The case of the DeepSeek-R1 model[90] exemplifies how optimising development efficiencies can achieve more with less hardware, a practice termed "Frugal AI"[60]. This demonstrated that a greater use of computation is not always synonymous with better results. Resources like Green Software Patterns[91] from the Green Software Foundation[92] can guide researchers in applying these energy-efficient practices throughout the model development cycle.

### R8: Choose and optimise hardware to reduce environmental impact

The optimisation and choice of hardware, relating to microchips and location, critically impacts AI sustainability. For on-premises computing, where possible researchers should select energy-efficient modern hardware, optimise energy settings and use chips optimised for specific tasks (e.g., GPUs for deep learning). With cloud computing, options are expanded to make use of data centres in green energy powered zones[55], scheduling workloads for optimal green energy availability. However, work with sensitive biological data and legal frameworks may limit such choices. For large scale AI projects, considering Top500 Green list facilities[93] is also recommended as they account for the full carbon life-cycle in their design and operation. Finally, leveraging available HPC optimisation support services (where accessible) like POP CoE in Europe[94] can also assist with reducing AI workload environmental impacts.

### R9: Measure and report AI model environmental impact

The life sciences community must begin to consistently measure and report AI model energy consumption in order to monitor their environmental impact. While no single metric is perfect due to measurement nuances and certain comparability limitations, researchers should aim to report core measures such as total model development runtime alongside hardware, energy consumption (e.g., kWh), and operational carbon emissions ($CO_2e$). Planning the collection of these measurements in advance is crucial, preferably leveraging active measurement tools (e.g., Carbontracker[55]) or where not possible rough predictive calculators (e.g., ML $CO_2$ Impact[95]). Measurements of final model inference impacts are also valuable to help researchers consider choice of AI models if planning reuse. Reporting of measures should ideally be made available in a publication's methods, or hosted on life science AI model evaluation platforms (e.g., OpenEBench[84]), targeted community low resource challenges such as MICCAI Fast and Low-resource (FLARE)[96] or dedicated leaderboards such as the Reinforcement Learning Energy Leaderboard[97].

**Box 3. Recommendations to aid AI model sustainability**

| Sustainability | | | |
|---|---|---|---|
| **No.** | **Recommendation** | **Explanation** | **OSAI Priority** |
| R7 | Implement Green AI model development techniques | Reduce energy consumption by implementing Green AI techniques (e.g., pruning, transfer learning) and adopting Frugal AI principles throughout the model development lifecycle, guided by established best practices. | Essential |
| R8 | Choose and optimise hardware to reduce environmental impact | Minimise environmental footprint by selecting energy-efficient hardware, optimising chip usage for specific tasks, and considering green energy sources or sustainably designed data centres for computations. | Required |
| R9 | Measure and report AI model environmental impact | Consistently measure and report key environmental impact metrics (e.g., energy consumption, carbon emissions, runtime) using suitable tooling, and disclose these figures in publications or on dedicated reporting platforms. | Recommended |

# Open and Sustainable AI: Ecosystem Mapping and Pathways

While best practices and recommendations can be impactful, they can face barriers to adoption by researchers resulting from a lack of clarity on how exactly to implement them. This can lead to a reduction in their overall application across the intended community. Therefore, we have gone beyond an isolated set of recommendations by creating a dedicated digital resource for OSAI to support practical implementation (https://osai.dome-ml.org/). The interactive resource details the OSAI recommendations, and contains a corresponding registry of over 300 relevant components of the AI ecosystem. With this list of components we have created a non-prescriptive mapping, connecting the OSAI relevant components that can help achieve our recommendations (Fig. 2). The mapping has inbuilt redundancy as it hosts many solutions for achieving the same recommendation, which can avoid situations such as the FAIR principles for AI models by Ravi et al. (2022)[98], wherein a key component for implementation, DLHub, is no longer available for use.

**Figure 2: Open and Sustainable AI: Recommendations & Ecosystem Components**

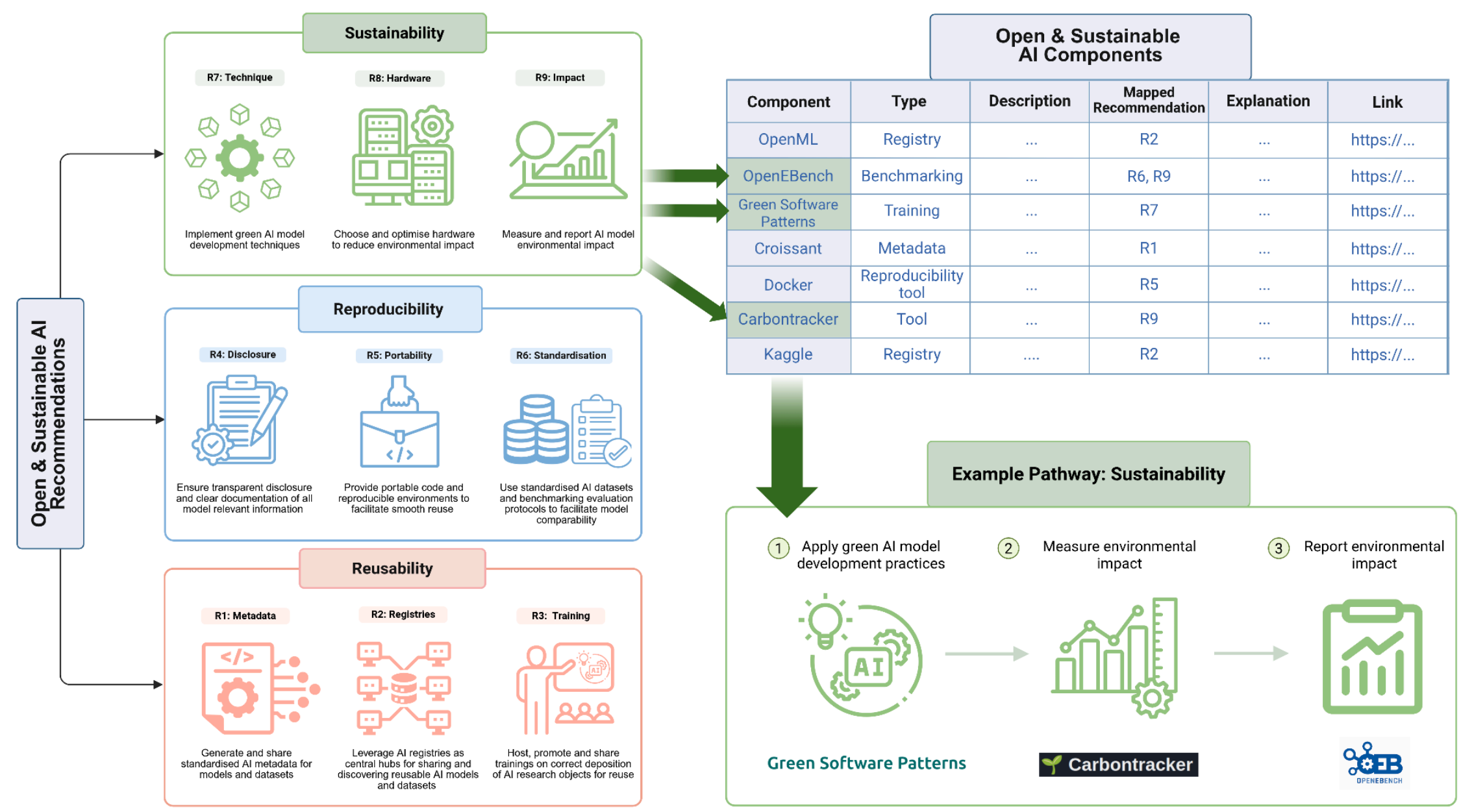

Open and Sustainable AI (OSAI) recommendations covering nine practical and actionable solutions addressing practical issues across the three challenge areas of reusability, reproducibility and environmental sustainability. These recommendations are directly mapped to the OSAI ecosystem components to support their implementation. The OSAI ecosystem components mapped offer flexibility and inbuilt redundancy with multiple solutions per recommendation across the full list. Both the OSAI recommendations and ecosystem components can further be chained into actionable pathways to best guide researchers to implement AI best practices, as shown with an example pathway covering environmentally sustainable model reporting.

We have designed the AI ecosystem components list to be maintainable by the community in an open GitHub (https://github.com/BioComputingUP/OSAI_ecosystem[99]. This method is directly modelled after proven community supported resource lists, such as within the Research Software Quality kit (RSQkit)[100]. In this way our landscaping builds upon the previous efforts shared in static publications, for example FAIR for AI by Huerta et al. (2023)[40]. This collaborative approach will support longevity and contemporary relevance of the components captured, made possible through ongoing community maintenance encouraged by prominent contributor credit and recognition.

The AI ecosystem components and OSAI recommendations mapping are designed to be an openly accessible resource for linking researchers to the many possible components enabling OSAI research practice. The ecosystem components captured are not solely restricted to OSAI, and also include important AI research platforms, tools and training, which expands its functionality to aid general AI research solution discoverability. We recognise the diversity across life science sub-domains and organisations, which may adopt varying AI guidance of narrower, wider or overlapping scope to those of our OSAI recommendations. Therefore, the captured AI ecosystem components are openly shared for interoperable reuse by other initiatives. For example they could be repurposed for mapping to existing life science AI research guidance frameworks, or hosted on other AI platforms to aid connecting researchers to relevant supports.

Fig. 2 highlights the OSAI recommendation mappings to the AI ecosystem components and how these can be further structured into guided OSAI implementation pathways. The OSAI web resource currently hosts three reference OSAI implementation pathways (https://osai.dome-ml.org/pathways) which cover reusability, reproducibility, and environmental sustainability. They chain together the OSAI recommendations with a narrower subset of recommended components to actionably achieve OSAI in a step-by-step manner. Long-term, we envision wider community adaptations of our OSAI reference implementation pathways into sub-domain specific pathways that highlight the most relevant AI ecosystem components in more targeted assemblies. We foresee Turing Way[75] modelled toolkit resources, which are designed to support and disseminate community co-created research best practices, as resources that could be leveraged for hosting these pathways. Several existing toolkits addressing data[12,101,102] or software[100] guidance could be considered for reuse to avoid proliferation of resources requiring large community and financial commitments to create, host and maintain them. We view additional guided pathways as the important next step towards supporting OSAI research practice across sub-domains and encourage global bodies working on mitigating challenges related to AI research come together to co-develop further pathways to guide AI best practice implementations.

## Emerging Challenges & Opportunities for OSAI

An emerging AI application in the life sciences is the use of AI agents and advanced agentic AI systems. AI agents can independently perform specific tasks of a narrow remit and can be further orchestrated into cohesive agentic AI systems to autonomously undertake diverse tasks with minimal human interaction[19]. Example AI agent applications include ontology maintenance (e.g., Uberon[103]) and spatial omics data analyses (e.g., OmicsNavigator[104]). More complex agentic AI systems like Biomni[105] have demonstrated the ability to perform biomedical research tasks ranging from generating wet-lab cloning protocols to analysing heterogeneous datasets for rare disease diagnosis. The foreseen growth and uptake of these impressive agentic AI resources raise additional concerns for ensuring OSAI compliant research methods.

Agentic AI systems introduce major OSAI concerns for both reproducibility[106] and environmental sustainability[107]. Regarding reproducibility, these systems operate across dynamic data sources, tools, and LLMs, which can lead to varied outcomes when tasks are executed across different periods due to underlying component changes such as LLM updates. In response, transparent documentation, thorough task execution logs, and open information sharing with an emphasis on Explainable AI are vital to ensure interpretable and reproducible agentic systems. Regarding environmental sustainability, the increased computational demands of agentic AI are a key concern. These systems can run for longer and therefore use significantly more energy than traditional LLM prompting. Additionally, the common need for repeated runs when tasks fail further increases resource usage. Therefore, leveraging OSAI recommendations such as the implementation of Green AI techniques and accountable resource usage reporting will be crucial for monitoring and reducing negative agentic AI impacts.

Another challenge for OSAI alignment is within the evolving frameworks of global policies like the EU AI Act[108], and the US' NAIRR[109] and related AI Action Plan[110]. While such AI frameworks offer high-level policy on areas of Responsible AI, most lack direct implementation guidance and are not yet closely connected to Open Science strategies to support reusable, reproducible and environmentally sustainable AI outputs. Addressing this gap at the policy level would enhance the uptake of AI best practices like OSAI in academia by feeding into compliance mandates from funders and institutions during their grant proposal assessments and researcher assessments respectively.

Currently, publications take priority in research assessment. To tackle this, global efforts like the Declaration on Research Assessment (DORA)[111] and the Coalition for Advancing Research Assessment (CoARA)[112] are reforming research assessment practices to elevate the evaluation of research assets beyond publications. Such frameworks can help introduce researcher incentives to implement AI best practices by rewarding excellence of non-traditional research artefacts, such as formal contribution towards career progression and recognition during funding proposal decisions in support of research best practice adherence.

Industry, in contrast, faces even starker barriers to adoption of Open Science and FAIR best practices. By design, companies must generate profit. They are therefore financially motivated and have less capacity or incentive to advance science for open exploitation in the same way as academia. This misalignment stemming from divergent incentives is a major concern for AI research asset outputs. Mitigation strategies such as stronger journal publication rules, and new policy and legislation could help to address this. Additionally, stipulations within academic collaborations to adhere to Open Science and FAIR principles could facilitate a closer alignment. This is important to tackle with the growing number of AI companies extending efforts into life science research such as Google Deepmind's AlphaFold[3] and Apple's SimpleFold[113]. A flagship example of these issues is the AlphaFold3[114] model, which was published without any available code. This led to heavy criticism[115] and necessitated a response[116] to address this.

Beyond these sectoral incentive challenges, a final cross-cutting consideration is the need to extend socio-ethical dimensions into AI research best practice frameworks. While OSAI does not cover these areas, future extensions could leverage collaborative learnings from successful initiatives tackling socio-ethical concerns in the area of sensitive data such as the GA4GH[117], Bridge2AI[8], and GDI Europe[118], which have effectively incorporated ELSI experts in the design of their outputs. Possible extensions and cross-walks of this work (or other frameworks) could be undertaken through cross-domain international working groups (e.g., via the RDA[16]) to facilitate this process. This will help ensure not only are AI research outputs open and environmentally sustainable but also ethical, responsible, and accountably designed.

# Conclusion

Our OSAI recommendations offer practical support for researchers, developed with close consideration for both existing life science and domain-agnostic solutions (e.g. from the RDA[41], MLCommons[24], Green Software Foundation[92]). They are designed to be complementary to comprehensive frameworks of wider AI scope, and directly focus on supporting research reform towards the production of OSAI research outputs. This work aims to foster a better understanding of OSAI practices and provides researchers with a navigable list of implementable solutions by consolidating the fragmented AI ecosystem components, and connecting them to actionable solutions.

While addressing the urgent challenges in areas of reusability, reproducibility, and sustainability, we recognise the limitations of this work. We acknowledge a primary focus on AI models and less coverage of the important area of achieving AI-ready data, but instead refer to the existing efforts and guidance in this space, such as Bridge2AI[8], the ODI's Framework for AI-ready data[43] and other supporting entries captured in our OSAI ecosystem. Furthermore, we do not directly address critical socio-ethical areas such as Equitable, Explainable, Trustworthy or full breadth of Responsible AI, given their complex, and variable requirements across regions. Instead we hope existing and future frameworks addressing these important areas can make use of our catalogue of AI ecosystem components, which is suitable to be repurposed for expanded resource mappings. We also recognise the initial landscaping of ecosystem components is not exhaustive, and will rely on open community contributions to ensure wider coverage, such as to incorporate emerging AI model and dataset deposition trainings as they become available to better support our OSAI recommendation R3 'Host, promote and share training and guidance on correct deposition of AI research objects for reuse'.

As a next step, the life sciences community must urgently fill the gap and collaboratively develop tailored AI best practice implementation pathways. Our work lays the foundations through expanding the availability of practical OSAI recommendations supported by a diverse range of useful resources and provision of three reference implementation pathways. We foresee additional guided pathways as the necessary next step, potentially hosted in a new or existing toolkit-based resource, to assist life science AI researchers in a targeted manner. Ultimately, fostering a truly OSAI landscape necessitates a concerted effort from all stakeholders, including publishers, funders, and policymakers. It will also be important to align community-driven efforts to the evolving policy landscape (e.g., EU AI Act[108], NAIRR[109]), and next integrate with global networks such as FARR-RCN[25] or large-scale projects like AI4EOSC[119].

By embracing our nine OSAI recommendations, leveraging the mapped ecosystem of AI components and reference implementation pathways, researchers can usher in a shift towards reusable, reproducible, and environmentally responsible AI becoming commonplace. This will ensure that life science AI outputs serve as reliable, and sustainable drivers for future life science breakthroughs serving society.

# Acknowledgments

This work has been supported by ELIXIR, the European infrastructure for life science data, through the ELIXIR Machine Learning Focus Group (Strategic Implementation Study 2023-MLstandards). Additional funding from: European Union through NextGenerationEU PNRR project ELIXIRxNextGenIT (IR0000010); Horizon Europe projects EVERSE (GA 101129744) and ELIXIR STEERS (GA 101131096); Funded by the European Union through the Marie Skłodowska-Curie projects AHEAD (grant agreement No 101183031) and IDPfun2 (grant agreement no. 101182949); Innovative Health Initiative Joint Undertaking SYNTHIA (grant agreement No 101172872); COST Action ML4NGP (CA21160), supported by COST (European Cooperation in Science and Technology); German Research Foundation (Deutsche Forschungsgemeinschaft - DFG) project number 460234259/NFDI for Data Science and Artificial Intelligence. Views and opinions expressed are those of the author(s) only and do not necessarily reflect those of the European Union or the European Research Executive Agency. Neither the European Union nor the granting authority can be held responsible for them.

# Author contributions

All authors contributed to the OSAI ecosystem and subsequent discussion as well as the initial draft of this manuscript. G.F. wrote the final draft with the help of the co-authors. All authors edited and refined the final manuscript. G.F., F.P. and S.C.E.T. initiated and coordinated the project.

# Competing interests

All authors declare no competing interests.

## Additional information

Correspondence should be addressed to Fotis Psomopoulos or Silvio C. E. Tosatto.

## Data availability

- **OSAI Ecosystem GitHub:** https://github.com/BioComputingUP/OSAI_ecosystem
- **OSAI Ecosystem Interactive Website:** https://osai.dome-ml.org/ai-ecosystem

## References

1. Walsh, I. *et al.* DOME: recommendations for supervised machine learning validation in biology. *Nat. Methods* **18**, 1122–1127 (2021).

2. Wilkinson, M. D. *et al.* The FAIR Guiding Principles for scientific data management and stewardship. *Sci. Data* **3**, 160018 (2016).

3. Jumper, J. *et al.* Highly accurate protein structure prediction with AlphaFold. *Nature* **596**, 583–589 (2021).

4. Luo, M. *et al.* Artificial intelligence for life sciences: A comprehensive guide and future trends. *Innov. Life* **2**, 100105 (2024).

5. Paysan-Lafosse, T. *et al.* The Pfam protein families database: embracing AI/ML. *Nucleic Acids Res.* **53**, D523–D534 (2025).

6. Varadi, M. *et al.* AlphaFold Protein Structure Database: massively expanding the structural coverage of protein-sequence space with high-accuracy models. *Nucleic Acids Res.* **50**, D439–D444 (2022).

7. Kapoor, S. & Narayanan, A. Leakage and the reproducibility crisis in machine-learning-based science. *Patterns* **4**, 100804 (2023).

8. Clark, T. *et al.* AI-readiness for Biomedical Data: Bridge2AI Recommendations. Preprint at https://doi.org/10.1101/2024.10.23.619844 (2024).

9. Tedersoo, L. *et al.* Data sharing practices and data availability upon request differ across scientific disciplines. *Sci. Data* **8**, 192 (2021).

10. Laurinavichyute, A., Yadav, H. & Vasishth, S. Share the code, not just the data: A case study of the reproducibility of articles published in the Journal of Memory and Language under the open data policy. *J. Mem. Lang.* **125**, 104332 (2022).

11. Fatima, N. *et al.* RDMkit: The Research Data Management Toolkit for Life Sciences. *Proc. Conf. Res. Data Infrastruct.* **1**, (2023).

12. Pistoia Alliance. The FAIR Toolkit for Life Science Industry. https://fairtoolkit.pistoiaalliance.org/.

13. Ouyang, W. *et al.* BioImage Model Zoo: A Community-Driven Resource for Accessible Deep Learning in BioImage Analysis. Preprint at https://doi.org/10.1101/2022.06.07.495102 (2022).

14. Avsec, Ž. *et al.* The Kipoi repository accelerates community exchange and reuse of predictive models for genomics. *Nat. Biotechnol.* **37**, 592–600 (2019).

15. Akhtar, M. *et al.* Croissant: A Metadata Format for ML-Ready Datasets. in *Proceedings of the Eighth Workshop on Data Management for End-to-End Machine Learning* 1–6 (2024). doi:10.1145/3650203.3663326.

16. Research Data Alliance. RDA FAIR for Machine Learning (FAIR4ML) Interest Group. https://www.rd-alliance.org/groups/fair-machine-learning-fair4ml-ig/activity/.

17. Beam, A. L., Manrai, A. K. & Ghassemi, M. Challenges to the Reproducibility of Machine Learning Models in Health Care. *JAMA* **323**, 305 (2020).

18. Unsal, S. *et al.* Learning functional properties of proteins with language models. *Nat. Mach. Intell.* **4**, 227–245 (2022).

19. Sapkota, R., Roumeliotis, K. I. & Karkee, M. AI Agents vs. Agentic AI: A Conceptual Taxonomy, Applications and Challenges. Preprint at https://doi.org/10.48550/ARXIV.2505.10468 (2025).

20. Schwartz, R., Dodge, J., Smith, N. A. & Etzioni, O. Green AI. *Commun. ACM* **63**, 54–63 (2020).

21. White, M. *et al.* The Model Openness Framework: Promoting Completeness and Openness for Reproducibility, Transparency, and Usability in Artificial Intelligence. Preprint at https://doi.org/10.48550/ARXIV.2403.13784 (2024).

22. FUTURE-AI: international consensus guideline for trustworthy and deployable artificial intelligence in healthcare. *BMJ* r340 (2025) doi:10.1136/bmj.r340.

23. Kapoor, S. *et al.* REFORMS: Consensus-based Recommendations for Machine-learning-based Science. *Sci. Adv.* **10**, eadk3452 (2024).

24. Machine Learning Commons. MLCommons - Better AI for Everyone. https://mlcommons.org/.

25. FAIR Advanced Research and Reproducibility (FARR) Research Coordination Network (RCN). *FARR RCN* https://www.farr-rcn.org.

26. Rai, A. Explainable AI: from black box to glass box. *J. Acad. Mark. Sci.* **48**, 137–141 (2020).

27. Afroogh, S., Akbari, A., Malone, E., Kargar, M. & Alambeigi, H. Trust in AI: progress, challenges, and future directions. *Humanit. Soc. Sci. Commun.* **11**, 1568 (2024).

28. Leslie, D. *Understanding Artificial Intelligence Ethics and Safety: A Guide for the Responsible Design and Implementation of AI Systems in the Public Sector.* https://zenodo.org/record/3240529 (2019).

29. Dignum, V. Responsible Artificial Intelligence -- from Principles to Practice. Preprint at https://doi.org/10.48550/arXiv.2205.10785 (2022).

30. Ahdritz, G. *et al.* OpenFold: retraining AlphaFold2 yields new insights into its learning mechanisms and capacity for generalization. *Nat. Methods* **21**, 1514–1524 (2024).

31. Collins, G. S. *et al.* TRIPOD+AI statement: updated guidance for reporting clinical prediction models that use regression or machine learning methods. *BMJ* e078378 (2024) doi:10.1136/bmj-2023-078378.

32. Schmied, C. *et al.* Community-developed checklists for publishing images and image analyses. *Nat. Methods* **21**, 170–181 (2024).

33. Liu, X. *et al.* Reporting guidelines for clinical trial reports for interventions involving artificial intelligence: the CONSORT-AI extension. *Nat. Med.* **26**, 1364–1374 (2020).

34. Cruz Rivera, S. *et al.* Guidelines for clinical trial protocols for interventions involving artificial intelligence: the SPIRIT-AI extension. *Nat. Med.* **26**, 1351–1363 (2020).

35. Kaggle. Kaggle: Your Machine Learning and Data Science Community. https://www.kaggle.com/.

36. Wolf, T. *et al.* HuggingFace's Transformers: State-of-the-art Natural Language Processing. Preprint at https://doi.org/10.48550/ARXIV.1910.03771 (2019).

37. Turon, G., Legese, A., Arora, D. & Duran-Frigola, M. Ersilia Model Hub: a repository of AI/ML models for infectious and neglected tropical diseases. Zenodo https://doi.org/10.5281/ZENODO.7274645 (2025).

38. European Organization For Nuclear Research & OpenAIRE. Zenodo: Research. Shared. Preprint at https://doi.org/10.25495/7GXK-RD71 (2013).

39. Leo, S. *et al.* Recording provenance of workflow runs with RO-Crate. *PLOS ONE* **19**, e0309210 (2024).

40. Huerta, E. A. *et al.* FAIR for AI: An interdisciplinary and international community building perspective. *Sci. Data* **10**, 487 (2023).

41. Castro, L. J. *et al.* FAIR4ML-schema. Preprint at https://doi.org/10.5281/ZENODO.14002310 (2024).

42. Pistoia Alliance Organisation Website. *Pistoia Alliance* https://www.pistoiaalliance.org/.

43. *A Framework for AI-Ready Data*. https://theodi.hacdn.io/media/documents/A_framework_for_AI-ready_data.pdf.

44. Scientific Computing World. Pistoia Alliance launches DataFAIRy to drive AI adoption. https://www.scientific-computing.com/news/pistoia-alliance-launches-datafairy-drive-ai-adoption.

45. What is Reproducibility in Artificial Intelligence and Machine Learning Research? https://arxiv.org/html/2407.10239v1.

46. Carter, R. E., Attia, Z. I., Lopez-Jimenez, F. & Friedman, P. A. Pragmatic considerations for fostering reproducible research in artificial intelligence. *Npj Digit. Med.* **2**, 42 (2019).

47. Tiwari, D. D. *et al.* BioModelsML: Building a FAIR and reproducible collection of machine learning models in life sciences and medicine for easy reuse. Preprint at https://doi.org/10.1101/2023.05.22.540599 (2023).

48. Merkel, D. Docker: lightweight Linux containers for consistent development and deployment. *Linux J* 2:2 (2014).

49. Anaconda, Inc. Conda. https://anaconda.org/anaconda/conda.

50. Di Tommaso, P. *et al.* Nextflow enables reproducible computational workflows. *Nat. Biotechnol.* **35**, 316–319 (2017).

51. Köster, J. & Rahmann, S. Snakemake—a scalable bioinformatics workflow engine. *Bioinformatics* **28**, 2520–2522 (2012).

52. The Galaxy Community *et al.* The Galaxy platform for accessible, reproducible, and collaborative data analyses: 2024 update. *Nucleic Acids Res.* **52**, W83–W94 (2024).

53. Heil, B. J. *et al.* Reproducibility standards for machine learning in the life sciences. *Nat. Methods* **18**, 1132–1135 (2021).

54. Bisong, E. Google Colaboratory. in *Building Machine Learning and Deep Learning Models on Google Cloud Platform* 59–64 (Apress, Berkeley, CA, 2019). doi:10.1007/978-1-4842-4470-8_7.

55. Anthony, L. F. W., Kanding, B. & Selvan, R. Carbontracker: Tracking and Predicting the Carbon Footprint of Training Deep Learning Models. Preprint at https://doi.org/10.48550/arXiv.2007.03051 (2020).

56. Ritchie, H. *et al.* Hardware and energy cost to train notable AI systems. *Our World in Data* https://ourworldindata.org/grapher/hardware-and-energy-cost-to-train-notable-ai-systems (2023).

57. Gailhofer, P. *et al.* The role of Artificial Intelligence in the European Green Deal. *Policy Dep. Econ. Sci. Qual. Life Policies Eur. Parliam.* (2023).

58. Bolón-Canedo, V., Morán-Fernández, L., Cancela, B. & Alonso-Betanzos, A. A review of green artificial intelligence: Towards a more sustainable future. *Neurocomputing* **599**, 128096 (2024).

59. EMBL. Sustainability: Reports and Resources. https://www.embl.org/about/info/sustainability/reports-resources/.

60. Yamada, T., Tanaka, H., Suzuki, S. & Brian, Z. Frugal Machine Learning: Making AI More Efficient, Accessible, and Sustainable. Preprint at https://doi.org/10.36227/techrxiv.173385981.11102720/v1 (2024).

61. Tornede, T. *et al.* Towards Green Automated Machine Learning: Status Quo and Future Directions. *J. Artif. Intell. Res.* **77**, 427–457 (2023).

62. Johnson, S. G., Simon, G. & Aliferis, C. Regulatory Aspects and Ethical Legal Societal Implications (ELSI). in *Artificial Intelligence and Machine Learning in Health Care and Medical Sciences* (eds Simon, G. J. & Aliferis, C.) 659–692 (Springer International Publishing, Cham, 2024). doi:10.1007/978-3-031-39355-6_16.

63. Jefferson, E. *et al.* GRAIMatter: Guidelines and Resources for AI Model Access from TrusTEd Research environments (GRAIMatter). *Int. J. Popul. Data Sci.* **7**, (2022).

64. European Commission. AI for Health: Evaluation of Applications & Datasets (AHEAD). *CORDIS* https://cordis.europa.eu/project/id/101183031.

65. European Commission. HORIZON Europe: ELIXIR-STEERS Project. *CORDIS* https://cordis.europa.eu/project/id/101131096 (2024).

66. SustAInML - Sustainable AI and Machine Learning. *SUSTAIN ML Project* https://sustainml.eu/ (2021).

67. Software Sustainability Institute. Green DiSC: a Digital Sustainability Certification. https://www.software.ac.uk/GreenDiSC.

68. Geoscience and Remote Sensing Society (GRSS), I. GeoCroissant- A Metadata Framework for Geospatial ML-ready Datasets. *GRSS-IEEE* https://www.grss-ieee.org/events/geocroissant-a-metadata-framework-for-geospatial-ml-ready-datasets/ (2024).

69. Mitchell, M. *et al.* Model Cards for Model Reporting. in *Proceedings of the Conference on Fairness, Accountability, and Transparency* 220–229 (ACM, Atlanta GA USA, 2019). doi:10.1145/3287560.3287596.

70. Pushkarna, M., Zaldivar, A. & Kjartansson, O. Data Cards: Purposeful and Transparent Dataset Documentation for Responsible AI. Preprint at https://doi.org/10.48550/ARXIV.2204.01075 (2022).

71. Dasoulas, I., Yang, D. & Dimou, A. MLSea: A Semantic Layer for Discoverable Machine Learning. in *The Semantic Web* (eds Meroño Peñuela, A. et al.) vol. 14665 178–198 (Springer Nature Switzerland, Cham, 2024).

72. SciLifeLab Data Centre. SciLifeLab: Funder requirements and FAIR ML models. https://serve.scilifelab.se/docs/model-serving/fair/

73. Van Geest, G., Haefliger, Y., Zahn-Zabal, M. & Palagi, P. M. Using Glittr.org to find, compare and re-use online materials for training and education. *PLOS ONE* **19**, e0308729 (2024).

74. Data Carpentry. Data Carpentry Lessons. https://datacarpentry.org/lessons/.

75. The Turing Way Community. The Turing Way: A handbook for reproducible, ethical and collaborative research. Zenodo https://doi.org/10.5281/ZENODO.15213042 (2025).

76. ONNX. ONNX - Open Neural Network Exchange. https://onnx.ai/.

77. Attafi, O. A. *et al.* DOME Registry: implementing community-wide recommendations for reporting supervised machine learning in biology. *GigaScience* **13**, giae094 (2024).

78. Kurtzer, G. M., Sochat, V. & Bauer, M. W. Singularity: Scientific containers for mobility of compute. *PLOS ONE* **12**, e0177459 (2017).

79. Docker. Docker Hub Container Image Library. https://hub.docker.com.

80. Yuen, D. *et al.* The Dockstore: enhancing a community platform for sharing reproducible and accessible computational protocols. *Nucleic Acids Res.* **49**, W624–W632 (2021).

81. Clyburne-Sherin, A., Fei, X. & Green, S. A. Computational Reproducibility via Containers in Psychology. *Meta-Psychol.* **3**, (2019).

82. Kryshtafovych, A., Schwede, T., Topf, M., Fidelis, K. & Moult, J. Critical assessment of methods of protein structure prediction ( CASP )—Round XV. *Proteins Struct. Funct. Bioinforma.* **91**, 1539–1549 (2023).

83. Xiong, Z. *et al.* Crowdsourced identification of multi-target kinase inhibitors for RET- and TAU- based disease: The Multi-Targeting Drug DREAM Challenge. *PLOS Comput. Biol.* **17**, e1009302 (2021).

84. Capella-Gutierrez, S. *et al.* Lessons Learned: Recommendations for Establishing Critical Periodic Scientific Benchmarking. Preprint at https://doi.org/10.1101/181677 (2017).

85. Ash, J. T. & Adams, R. P. On Warm-Starting Neural Network Training. https://doi.org/10.48550/ARXIV.1910.08475 (2019) doi:10.48550/ARXIV.1910.08475.

86. Tmamna, J. *et al.* Pruning Deep Neural Networks for Green Energy-Efficient Models: A Survey. *Cogn. Comput.* **16**, 2931–2952 (2024).

87. Krishnan, S. & Faust, A. Quantization for Fast and Environmentally Sustainable Reinforcement Learning. https://research.google/blog/quantization-for-fast-and-environmentally-sustainable-reinforcement-learning/ (2021).

88. Yuan, Y. *et al.* The Impact of Knowledge Distillation on the Energy Consumption and Runtime Efficiency of NLP Models. in *Proceedings of the IEEE/ACM 3rd International Conference on AI Engineering - Software Engineering for AI* 129–133 (ACM, Lisbon Portugal, 2024). doi:10.1145/3644815.3644966.

89. Tabbakh, A. *et al.* Towards sustainable AI: a comprehensive framework for Green AI. *Discov. Sustain.* **5**, 408 (2024).

90. DeepSeek-AI *et al.* DeepSeek-R1: Incentivizing Reasoning Capability in LLMs via Reinforcement Learning. Preprint at https://doi.org/10.48550/arXiv.2501.12948 (2025).

91. Green Software Foundation. Green Software Patterns. https://patterns.greensoftware.foundation//.

92. Green Software Foundation. Green Software Foundation. https://greensoftware.foundation/.

93. TOP500.org. Green500 List - November 2023. https://top500.org/lists/green500/2023/11/.

94. Performance Optimisation and Productivity | A Centre of Excellence in HPC. https://pop-coe.eu/.

95. Schmidt, V., Luccioni, A., Lacoste, A. & Dandres, T. Machine Learning CO2 Impact Calculator. https://mlco2.github.io/impact.

96. Official Repository of MICCAI FLARE Challenges. https://github.com/JunMa11/FLARE.

97. Henderson, P. *et al.* Towards the systematic reporting of the energy and carbon footprints of machine learning. *J Mach Learn Res* **21**, 248:10039-248:10081 (2020).

98. Ravi, N. *et al.* FAIR principles for AI models with a practical application for accelerated high energy diffraction microscopy. *Sci. Data* **9**, 657 (2022).

99. Farrell, G. OSAI Ecosystem Components Data. Zenodo https://doi.org/10.5281/ZENODO.15391273 (2025).

100. RSQKit Community. Research Software Quality Kit (RSQKit). Zenodo https://doi.org/10.5281/ZENODO.14923572 (2025).

101. D'Anna, F. *et al.* A research data management (RDM) community for ELIXIR. *F1000Research* **13**, 230 (2024).

102. BY-COVID. Infectious Diseases Toolkit (IDTk). https://www.infectious-diseases-toolkit.org/about/.

103. Mungall, C. Open knowledge bases in the age of generative AI. Preprint at https://doi.org/10.7490/F1000RESEARCH.1120248.1 (2025).

104. Yiyao, L. *et al.* OmicsNavigator: an LLM-driven multi-agent system for autonomous zero-shot biological analysis in spatial omics. Preprint at https://doi.org/10.1101/2025.07.21.665821 (2025).

105. Huang, K. *et al.* Biomni: A General-Purpose Biomedical AI Agent. Preprint at https://doi.org/10.1101/2025.05.30.656746 (2025).

106. Wei, J. *et al.* From AI for Science to Agentic Science: A Survey on Autonomous Scientific Discovery. Preprint at https://doi.org/10.48550/ARXIV.2508.14111 (2025).

107. Kim, J., Shin, B., Chung, J. & Rhu, M. The Cost of Dynamic Reasoning: Demystifying AI Agents and Test-Time Scaling from an AI Infrastructure Perspective. Preprint at https://doi.org/10.48550/ARXIV.2506.04301 (2025).

108. European Commission. The EU AI Act. https://digital-strategy.ec.europa.eu/en/policies/regulatory-framework-ai (2024).

109. National Science Foundation. National Artificial Intelligence Research Resource (NAIRR) Pilot. https://www.nsf.gov/focus-areas/artificial-intelligence/nairr (2024).

110. The White House. *America's AI Action Plan*. https://www.whitehouse.gov/wp-content/uploads/2025/07/Americas-AI-Action-Plan.pdf

111. *Declaration on Research Assessment (DORA)*. https://sfdora.org/about-dora/.

112. *CoARA*. https://coara.org/.

113. Wang, Y., Lu, J., Jaitly, N., Susskind, J. & Bautista, M. A. SimpleFold: Folding Proteins is Simpler than You Think. Preprint at https://doi.org/10.48550/ARXIV.2509.18480 (2025).

114. Abramson, J. *et al.* Accurate structure prediction of biomolecular interactions with AlphaFold 3. *Nature* **630**, 493–500 (2024).

115. AlphaFold3 — why did Nature publish it without its code? *Nature* **629**, 728–728 (2024).

116. Callaway, E. AI protein-prediction tool AlphaFold3 is now more open. *Nature* **635**, 531–532 (2024).

117. Global Alliance for Genomics & Health (GA4GH). https://www.ga4gh.org/.

118. Pascucci, E. *et al.* Progressing towards personalised medicine: the Genomic Data Infrastructure (GDI) project. *Eur. J. Public Health* **34**, ckae144.1956 (2024).

119. Sáinz-Pardo Díaz, J. *et al.* Making Federated Learning Accessible to Scientists: The AI4EOSC Approach. in *Proceedings of the 2024 ACM Workshop on Information Hiding and Multimedia Security* 253–264 (Association for Computing Machinery, New York, NY, USA, 2024). doi:10.1145/3658664.3659642.

**Barriers to Open & Sustainable AI**

Reusability

Reproducibility

Sustainability

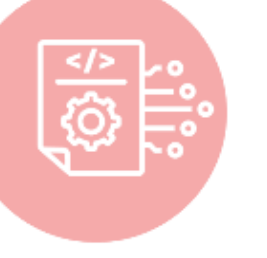

Low use of AI model and dataset specific metadata
**e.g.** FAIR4ML/Croissant

Low deposition of AI code, models and datasets in dedicated registries
**e.g.** Kaggle/Ersilia Model Hub

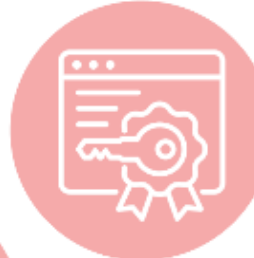

Absent or restrictive licensing for datasets and model code
**e.g.** Code repository has no license

Models not reusable, requiring more energy to train new ones
**e.g.** No model code shared

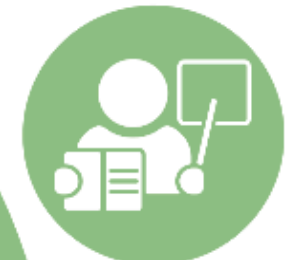

Inadequate mandates to monitor environmental impact
**e.g.** From funders

Insufficient spaces to report environmental impact measures
**e.g.** Leaderboards

Low reporting of environmental impacts even when measured
**e.g.** In publications

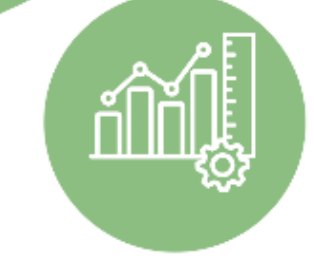

Too many environmental impact measures to choose from
**e.g.** Compute hours/$CO_2e$

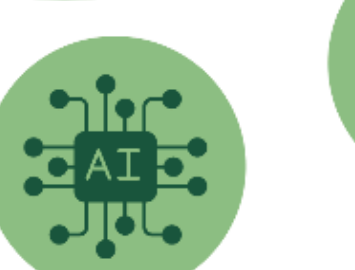

Low use of green AI approaches
**e.g.** Model distillation

Non-reproducible AI models leading to wasted compute & researcher resources
**e.g.** Non-portable code

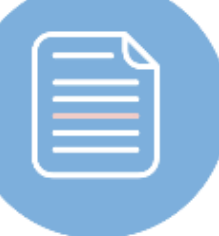

Inadequate sharing of software elements enabling reproducibility
**e.g.** Low Dockstore use

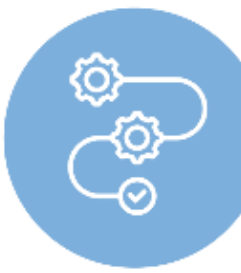

Low use of workflow management systems
**e.g.** Nextflow/Snakemake

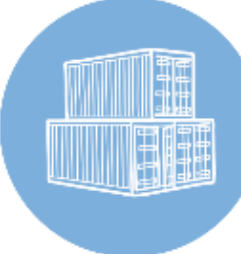

Low use of containers
**e.g.** Docker

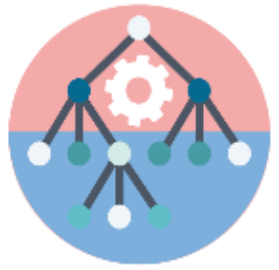

Low use of software environment controls
**e.g.** Conda

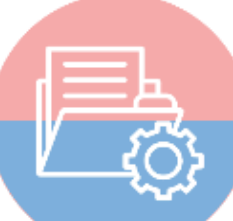

Low deposition of vital configuration files & reproducibility components
**e.g.** Containers

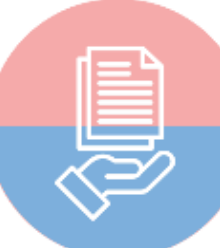

Low journal mandates for AI best practices
**e.g.** DOME/REFORMS

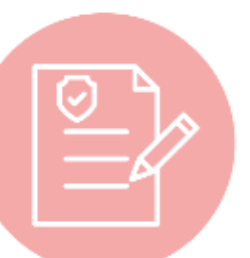

Poor AI methods disclosure
**e.g.** Dataset provenance

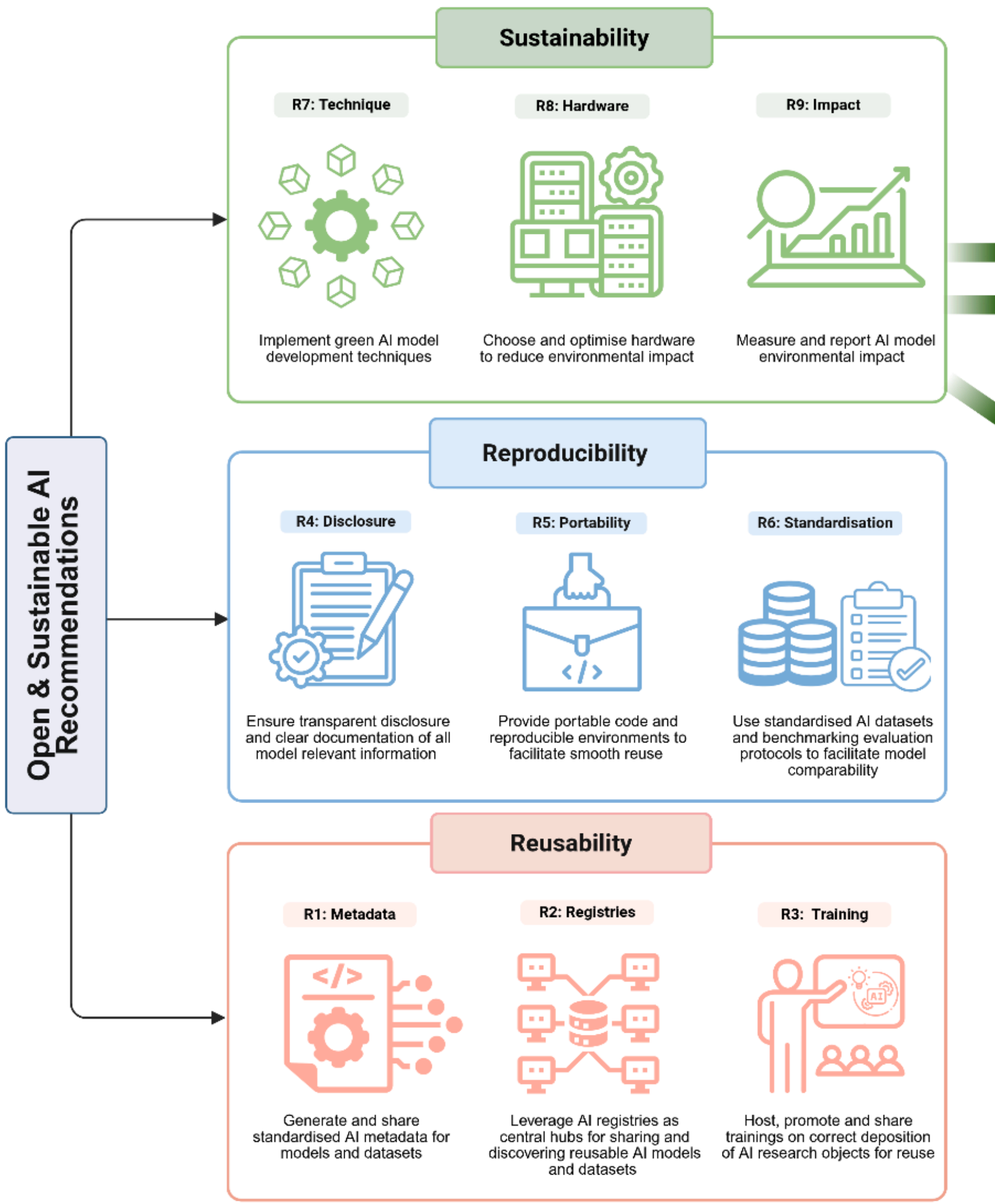

## Open & Sustainable AI Components

| Component | Type | Description | Mapped Recommendation | Explanation | Link |
|---|---|---|---|---|---|
| OpenML | Registry | ... | R2 | ... | https://... |
| OpenEBench | Benchmarking | ... | R6, R9 | ... | https://... |
| Green Software Patterns | Training | ... | R7 | ... | https://... |
| Croissant | Metadata | ... | R1 | ... | https://... |
| Docker | Reproducibility tool | ... | R5 | ... | https://... |
| Carbontracker | Tool | ... | R9 | ... | https://... |
| Kaggle | Registry | .... | R2 | ... | https://... |

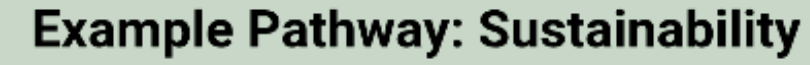

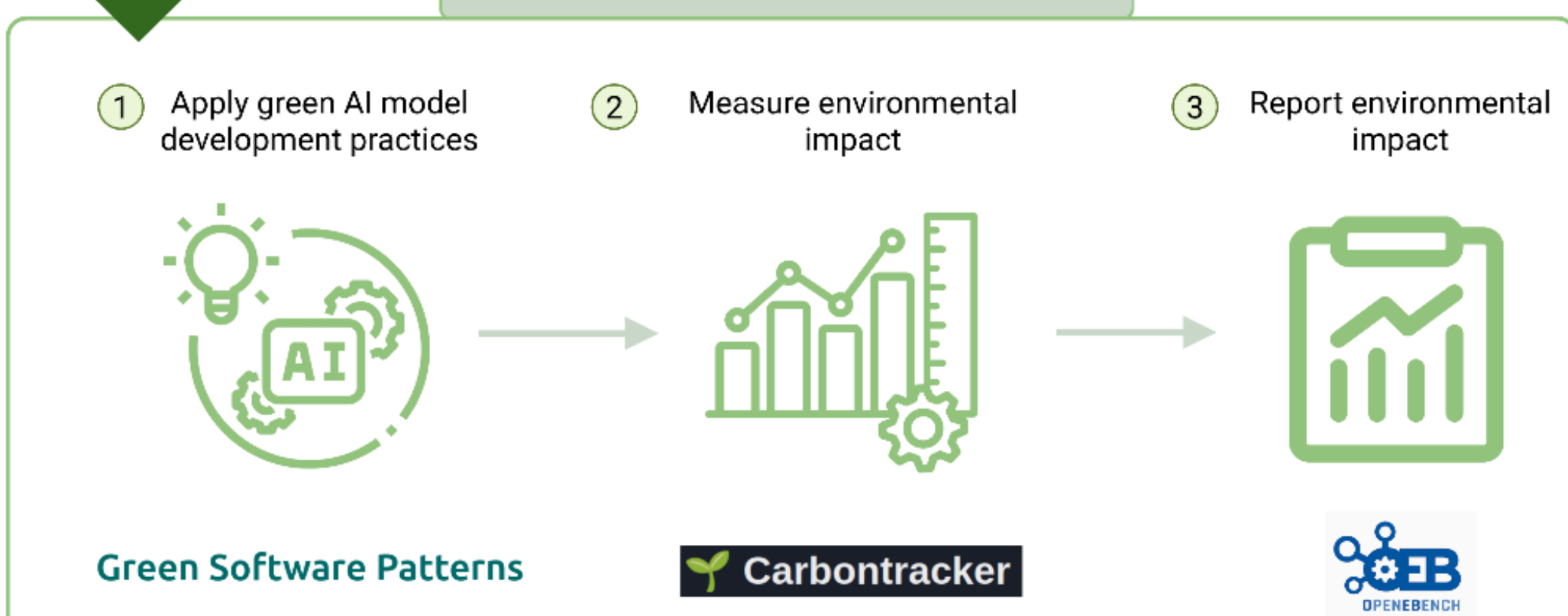